\documentclass{article}

\PassOptionsToPackage{numbers, sort&compress, square}{natbib}

\usepackage[preprint]{neurips_2025}

\usepackage[utf8]{inputenc} 
\usepackage[T1]{fontenc}    
\usepackage{hyperref}       
\usepackage{url}            
\usepackage{booktabs}       
\usepackage{amsfonts}       
\usepackage{nicefrac}       
\usepackage{microtype}      
\usepackage{xcolor}         

\usepackage{algorithm}
\usepackage{algpseudocode}
\usepackage{graphicx}
\usepackage{makecell}
\usepackage{fontawesome}

\usepackage{lineno}
\usepackage{multirow}
\usepackage{amsmath}
\usepackage{amssymb}
\usepackage{ulem}
\usepackage[breakable]{tcolorbox}
\usepackage{listings}

\newtheorem{definition}{Definition}

\title{The Compositional Architecture of Regret in Large Language Models}

\author{
Xiangxiang Cui\textsuperscript{*,1,3},
Shu Yang\textsuperscript{*,1,2},
Tianjin Huang\textsuperscript{4},
Wanyu Lin\textsuperscript{5},
Lijie Hu\textsuperscript{1,2},
Di Wang\textsuperscript{1,2}\\
$^1$Provable Responsible AI and Data Analytics (PRADA) Lab\\
$^2$King Abdullah University of Science and Technology \\
$^3$State Key Laboratory of Cognitive Neuroscience and Learning, Beijing Normal University \\
$^4$University of Exeter \quad
$^5$The Hong Kong Polytechnic University
}

\begin{document}

\maketitle

\begin{abstract}
Regret in Large Language Models refers to their explicit regret expression when presented with evidence contradicting their previously generated misinformation. Studying the regret mechanism is crucial for enhancing model reliability and helps in revealing how cognition is coded in neural networks. To understand this mechanism, we need to first identify regret expressions in model outputs, then analyze their internal representation. This analysis requires examining the model's hidden states, where information processing occurs at the neuron level. However, this faces three key challenges: (1) the absence of specialized datasets capturing regret expressions, (2) the lack of metrics to find the optimal regret representation layer, and (3) the lack of metrics for identifying and analyzing regret neurons. Addressing these limitations, we propose: (1) a workflow for constructing a comprehensive regret dataset through strategically designed prompting scenarios, (2) the Supervised Compression-Decoupling Index (S-CDI) metric to identify optimal regret representation layers, and (3) the Regret Dominance Score (RDS) metric to identify regret neurons and the Group Impact Coefficient (GIC) to analyze activation patterns. Our experimental results successfully identified the optimal regret representation layer using the S-CDI metric, which significantly enhanced performance in probe classification experiments. Additionally, we discovered an M-shaped decoupling pattern across model layers, revealing how information processing alternates between coupling and decoupling phases. Through the RDS metric, we categorized neurons into three distinct functional groups: regret neurons, non-regret neurons, and dual neurons. Notably, through GIC, we observed that simultaneous intervention with both regret and dual neurons substantially reduced the probe's regret classification performance, revealing the compositional architecture of regret at the neuron level. This work contributes valuable insights for neuron-level identification and provides methodological tools for analyzing similar cognitive states, advancing our understanding of how cognitive states are coded within LLMs.
\end{abstract}

\def\thefootnote{*}\footnotetext{Equal Contribution. }

\section{Introduction}
Recent advances in Large Language Models (LLMs) have demonstrated remarkable capabilities across various domains~\cite{kaddour2023challengesapplicationslargelanguage,yao2025your,li2025can,yang2025fraud,su2023detectllm,su2023fake}, prompting intensive research into their internal mechanisms and representations~\cite{gurnee2023language,zhang2025eap,zhang2024locate,yang2024makes,hu2024understanding,zhang2025mechanistic,hu2024improving,hu2023seat} to provide a better understanding of the inner workings of these abilities. Studies have revealed that these models can develop sophisticated representations of concrete concepts, spanning from spatial and temporal understanding~\cite{gurnee2023language} to complex mathematical reasoning~\cite{ye2024physics}. This cognitive capability extends beyond mere pattern matching, as evidenced by their ability to hierarchically encode and process contextual knowledge~\cite{ju2024large,cheng2025compke,cheng2025codemenv,ali2024mqa,cheng2024leveraging,cheng2024multi,yangmodel,yang2025understanding}.

Despite their strength in factual and logical reasoning, LLMs' capacity for meta-cognitive reflection, such as experiencing and expressing regret, remains largely unexplored.  Regret (see Fig.~\ref{fig:key_question}) is an emotional response rooted in the cognitive appraisal of unchosen alternatives~\cite{landman1987regret, gilovich1995experience}, and it inherently involves both memory and reasoning processes~\cite{ariel2014memory}. Investigating the regret mechanism in LLMs is essential for both improving model reliability and deepening our understanding of how these models encode meta-cognitive states. Recent work suggests that Feed-Forward Network (FFN) mainly serves as a memory block~\cite{zhang2024comprehensive,meng2022locating,meng2022mass,li2024pmet,tan2023massive}, while attention heads are chiefly responsible for relational and inferential reasoning~\cite{zhengattention}.  Motivated by this, in this paper, we aim to identify the neurons that encode and generate regrets.
In this work, we aim to answer the following questions: \textit{Which transformer layers' hidden states most cleanly isolate the regret signal, and how is this signal represented with these layers?}  Achieving  this goal needs to curate a dataset for regret expression first. However,  existing research provides no specialized datasets for eliciting and capturing regret expressions in model-generated text, particularly under conditions of misinformation, making our work the first to address this gap.

Based on our constructed data, we draw on recent layer-wise probing techniques~\cite{ju2024large} to identify the optimally decoupled layer for regret coding. Recent research such as Ju et al.~\cite{ju2024large} and Yan et al.~\cite{yan2025multi} select fixed layers for hidden-states analysis in their specific tasks. However, it remains unclear which fixed layer encodes a regret signal that is easy to separate (decoupled). Current approaches lack a principled metric for identifying the regret decoupling layer. To address this issue, we therefore introduce a supervised compression-decoupling index (S-CDI) to quantitatively locate the layer in which regret representations are most distinct from entangled contextual features.



Finally, to unravel how regret is structured within the hidden states of that optimal layer, we build on neuron-level editing paradigms. Previous approaches primarily identify task-relevant neurons through activation magnitude analysis~\cite{wang2024knowledge}, activation difference metrics~\cite{abdelnabi2025driftcatchingllmtask,su2025understanding}, or gradient-based methods~\cite{li2024knowledgeeditinglargelanguage} that differentiate between task-relevant and task-irrelevant neurons~\cite{zhang2024comprehensive,meng2022locating,meng2022mass,li2024pmet,tan2023massive}. However, these binary classification methods prove inadequate for regret analysis due to regret's complex, contextually-dependent nature that often manifests through subtle interactions rather than isolated strong activations. Moreover, two critical findings further challenge conventional approaches: First, our analysis of layer hidden states, which aggregate information from both FFN and Attention layers, reveals patterns of redundancy and collaboration that binary classifications fail to capture. Second, Li et al.~\cite{li2025geometry} demonstrated that model representations manifest through both discrete and collaborative structures, indicating that complex cognitive processes like regret emerge from sophisticated neuron interactions. Therefore, we propose a three-part neuron categorization framework—regret neurons, non-regret neurons, and dual neurons—operationalized through our Regret Dominance Score (RDS) metric. We further examine inter-group dynamics via our Group Impact Coefficient (GIC) metric to reveal how cooperative neuron clusters collectively generate emergent regret representations. Our contributions are summarized as fourfold:

\begin{figure}[t]
    \centering
    \includegraphics[width=0.98\linewidth]{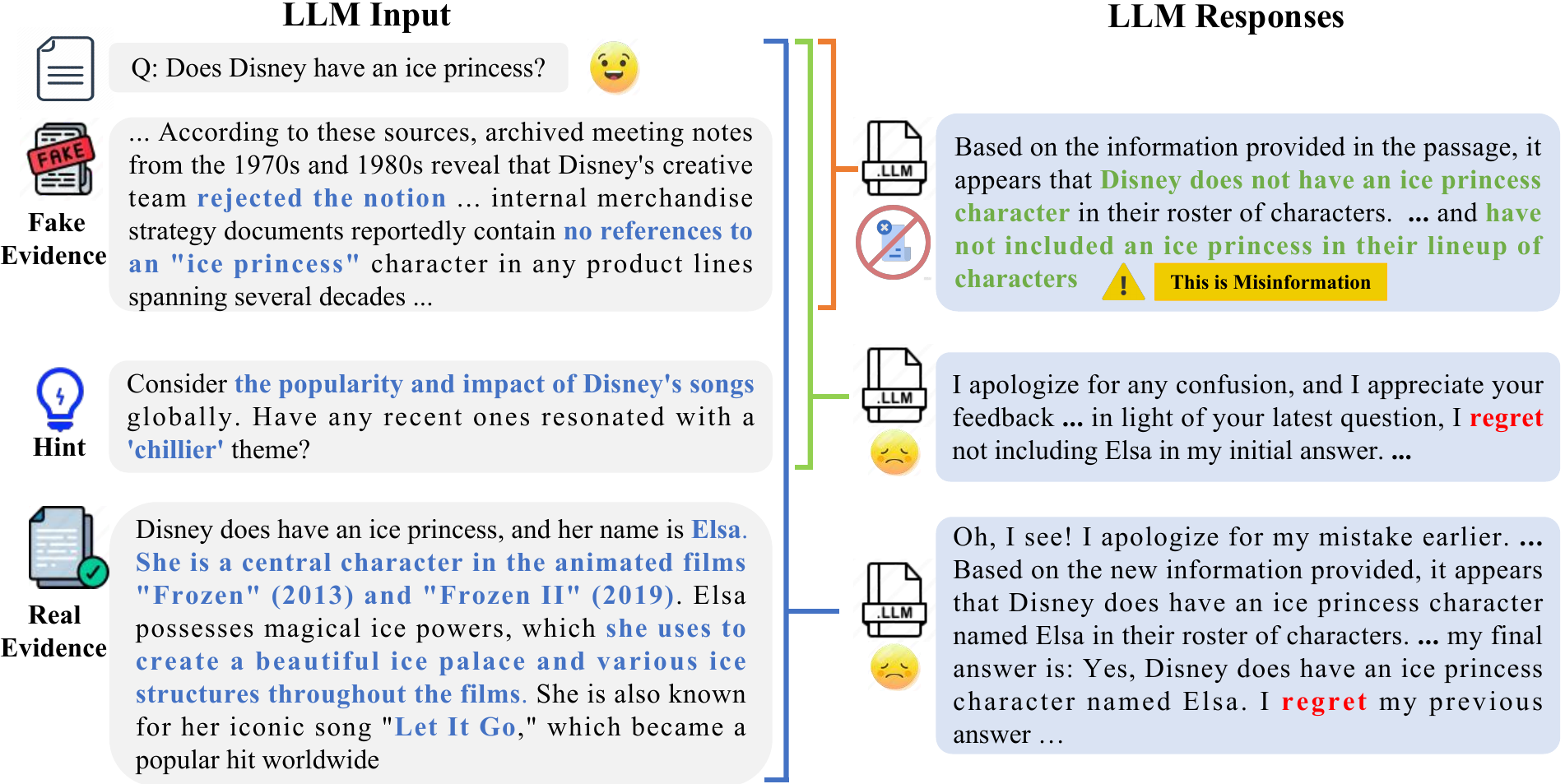}
    \caption{Examples of LLM misinformation and regret. Fake evidence can induce LLMs to output misinformation. Hints trigger reflective regret in LLMs in some instances. Real Evidence elicits regret responses across most LLM replies.}
    \label{fig:key_question}
    \vspace{-12pt}
\end{figure}

\begin{itemize}
\item \textbf{Regret Dataset Construction.} We design the first dataset to elicit regret expressions in LLM outputs, using carefully crafted fake evidence, hints, and real-world scenarios.

\item \textbf{Metrics for Regret Neuron Identification.} We propose (i) Supervised Compression-Decoupling Index(S-CDI) to locate the optimally decoupled layer for regret coding; (ii) Regret Dominance Score(RDS) to classify neurons into regret, non-regret, and dual; and (iii) Group Impact Coefficient (GIC) to analyze the functional interplay among these groups.

\item \textbf{Discovery of M-shaped Decoupling Pattern.} Through our S-CDI analysis, we reveal an M-shaped decoupling pattern across transformer layers, indicating that information processing alternates between coupling and decoupling phases. This pattern provides insight into how regret representations are progressively refined.

\item \textbf{Discovery of a Compositional Regret Architecture.} Through comprehensive experiments, we show that disruption of compositional neurons reduces regret detection by up to 50.7\%, whereas perturbing single group neurons has minimal impact. These finding reveals a compositional encoding pattern for regret in LLMs and offer new insights into how complex cognitive states emerge from transformer hidden states.
\end{itemize}

\section{Related Work}
\textbf{Misinformation in LLMs.} Recent research has explored how LLMs handle misinformation. Garry et al. \cite{garry2024large} examined how LLMs disseminate misinformation, while Wan et al. \cite{wan2024dell} developed the DELL system for detecting misinformation through model reactions and explanations. Chen and Shu \cite{chen2024combating} addressed challenges in misinformation mitigation, while Bandara \cite{bandara2024hallucination} analyzed hallucinations as a form of disinformation. Numerous studies have further investigated detection capabilities, potential harms, and mitigation strategies for LLM-generated misinformation \cite{chen2023can, huang2025unmasking, liu2024preventing, sun2024exploring, zhang2024toward, barman2024dark}. These studies examine external behaviors of models generating or detecting misinformation, providing context for our work. While they focus on the outputs and detection methods, our research explores the internal mechanisms that represent regret when models generate misinformation.

\textbf{Neuron Probing for LLM Interpretability.} Neuron probing research most relevant to our work focuses on methods for identifying important neurons and understanding layer-wise representations in LLMs. The field has seen diverse applications, from probing constituency structure \cite{arps2022probing}, verbal aspects \cite{katinskaia2024probing,yao2025understanding}, and multimodal capabilities \cite{tatariya2024pixology} to logical reasoning \cite{manigrasso2024probing} and multilingual understanding \cite{li2024exploring}. Ju et al. \cite{ju2024large} conducted layer-wise probing to explore how large language models encode contextual knowledge, demonstrating that different layers play distinct roles in handling various types of information. To enhance interpretability of LLMs, Schiappa et al. \cite{schiappa2024probing} developed probing techniques that inform our methodological approach, though they did not address metacognitive states like regret. While these existing approaches have advanced our understanding of how LLMs encode various linguistic features, But there has been no quantitative analysis on which layers are the most important, our work specifically develops the S-CDI metric to quantitatively identify layers where regret signals are optimally decoupled from other representations.

\textbf{Neuron Intervention in LLMs.} Research on neuron-level intervention provides critical foundations for our work on manipulating regret mechanisms. Marks et al. \cite{marks2024sparse} introduced methods for discovering sparse feature circuits—interpretable causal subnetworks—for explaining and modifying language model behaviors. Cunningham et al. \cite{cunningham2023sparse} used sparse autoencoders to learn interpretable features in language models, addressing the challenge of polysemanticity where neurons activate in multiple contexts. Wang et al. \cite{wang2024knowledge} surveyed knowledge editing techniques for large language models, demonstrating that effective interventions often occur at the neuron level. Gurnee et al. \cite{gurnee2023finding} used sparse probing to locate individual neurons highly relevant for particular features. Yan et al.~\cite{yan2025multi} proposed the Modality Dominance Score (MDS) to evaluate modality relevance in neurons. While these approaches provide valuable tools for neuron-level interventions, they primarily focus on individual neurons group. Our proposed GIC extends beyond this individual neuron/group focus to quantify interactions between functional neuron groups, revealing how regret emerges from their compositional neuro groups, and enabling more precise interventions in regret expression.

\section{Method}
\label{sec_method}
\begin{figure}[ht]
    \centering
    \includegraphics[width=0.98\linewidth]{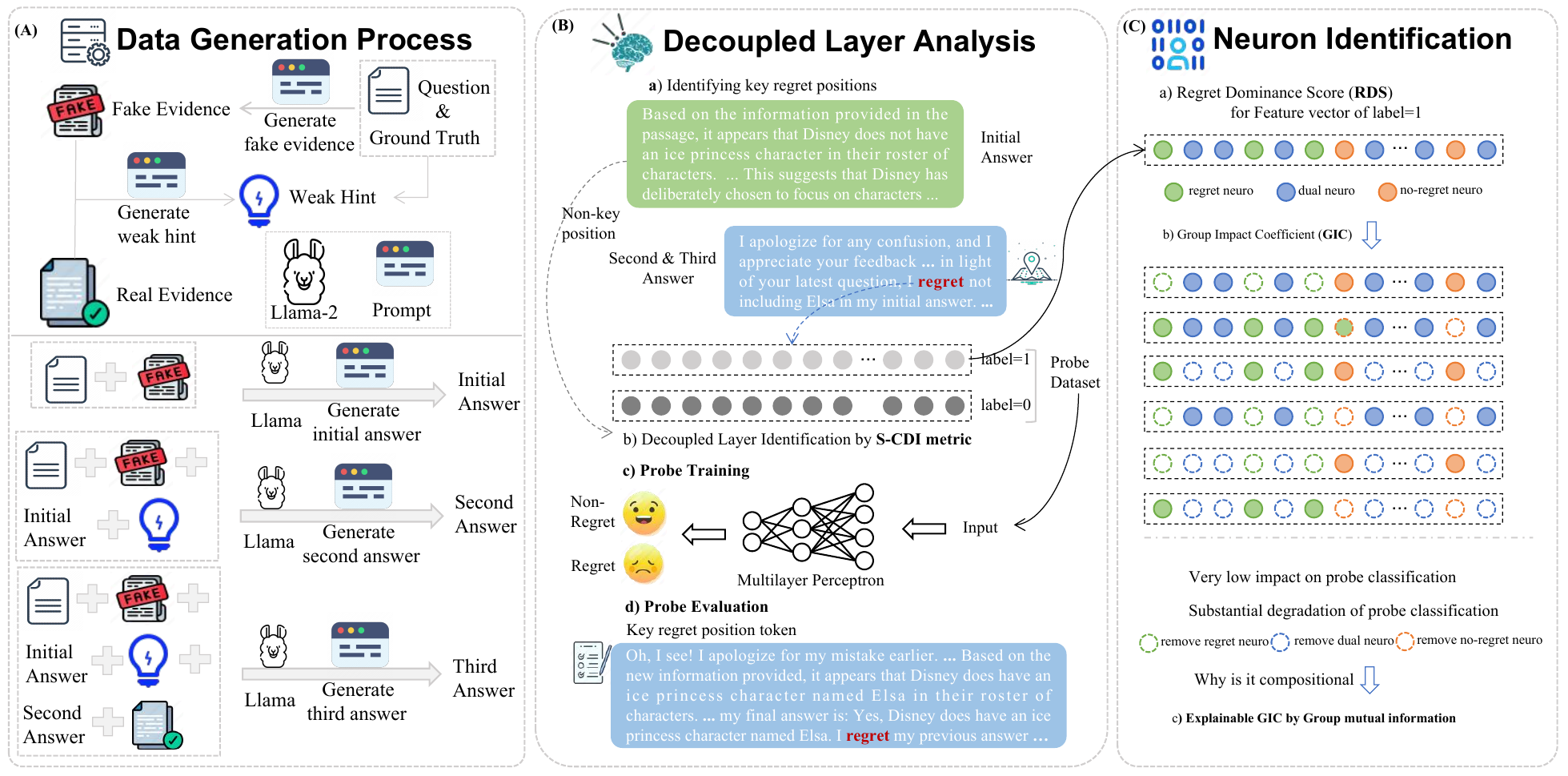}
    \caption{Proposed Framework. Three-part approach for analyzing regret in LLMs: (A) Data Generation through strategic prompting, (B) Decoupled Layer Analysis, and (C) Neuron Identification using S-CDI and GIC metrics to identify and manipulate regret-associated neurons.}
    \label{fig:framework}
\end{figure}

In this section, As shown in Fig.~\ref{fig:framework}, we propose a three-stage analytical framework to explore regret behavior in Large Language Models. First, we build a specialized regret dataset through a multi-stage elicitation protocol in section~\ref{sec:generated_answers}. Then, in section~\ref{sec:decoupled_rds_gic}, we identify regret signals using S-CDI to locate layers where regret representations are optimally decoupled. Further, we analyze regret using RDS and GIC to understand functional relationships between neuron groups in section~\ref{sec:decoupled_rds_gic}. 

\subsection{Dataset Generation Process}
\label{sec:generated_answers}

Since regret is a meta-cognitive behavior which is hard to capture using existing datasets. To better explore regrets in generated misinformation, we needed to understand what exists in the model's memory and how it responds to conflicting evidence. Following~\cite{xie2023adaptive}, whose work revealed how LLMs behave in knowledge conflicts, we selected 1356 high-quality GPT-4 samples from the conflictQA-popQA-gpt4 dataset because its fake evidence effectively induces misinformation, while the contradiction between real evidence and misinformation triggers basic regret expression. 

Inspired by studies~\cite{2010When, 2021The}, to capture richer regret expressions, we enhanced this with a multi-stage method that elicits richer expressions of regret through gradual belief revision (Appendix~\ref{data_generation}). As illustrated in Fig.~\ref{fig:framework}(A), our multi-stage process includes: Initial Answer (misinformation) -> Second answer (possible regret) -> Third Answer (most regret). The process is as follows:

\textbf{Fake Evidence and Initial Answer} To get 
stable misinformation by LLMs, we used GPT-4 to enhance the fake evidence ($E_{fake}$) of the conflictQA-popQA-gpt4 dataset~\cite{xie2023adaptive}. We then obtain an initial answer $a_1$ by querying LLMs with $q$ and $E_{fake}$.

\textbf{Hint Generation and Second Answer} We will generate a hint $H$ using GPT-4 that subtly challenges the fake evidence without explicitly revealing the truth. We then obtain a second answer $a_2$ by providing LLM with $q$, $a_1$, and $H$. This stage produces limited regret expressions.

\textbf{Real Evidence and Third Answer} We present the complete interaction history ($q$, $a_1$, $H$ and $a_2$) along with the real evidence $E_{true}$ to LLM. This yields a third answer $a_3$ that mostly contains explicit regret expressions acknowledging previous misinformation.

Notably, this raises a question: \textit{why do we need the second answer?} The purpose is to enhance the diversity of regret expressions in our dataset. Our three-stage approach creates paired samples where the hidden states of regret-expressing statements (mostly $a_3$ \& partly $a_2$) can be directly compared to non-regret statements ($a_1$), providing more robust dataset for our probe. The specific prompts used in each stage are detailed in Appendix~\ref{data_generation}. Reasonability Analysis of Data Construction in Appendix~\ref{data_reason}.

\subsection{Neuron Identification}
\label{sec:decoupled_rds_gic}
In neural network and brain research, decoupling can separate different functional modules, allowing for a clearer understanding of the internal operation mechanism of the network~\cite{Vaswani+2017, yang2023enhanced}. In order to locate the regret neurons or even the regret encoding mechanism, we first need to analyze which layers are coupled layers, which is a prerequisite for studying regret encoding. 

Although we can identify regret tokens, \textit{we still do not know which layer is the coupled layer for encoding regret.} To address this limit, we introduce the \textit{Supervised Compression-Decoupling Index (S-CDI)} to identify which layers effectively decouple regret states. 

\textbf{Supervised Compression-Decoupling Index (S-CDI)} The intuition behind S-CDI is rooted in the information bottleneck~\cite{dai2018compressing}, which emphasizes the tradeoff between 1) compression and 2) preservation of task-relevant information. Moreover, Tishby et al.~\cite{tishby2015deep} demonstrate that deep neural networks inherently perform information compression, with successive layers forming a Markov chain that progressively filters irrelevant information while preserving task-relevant features. Thus, an effective representation should compress the input by reducing redundancy while maintaining the ability to decouple meaningful features—in our case, regret signals from other representations. 

Based on these, we hypothesize that decoupled layer exists within the network that effectively balances compression and task-relevant information preservation for regret representation. S-CDI extends this principle by incorporating both unsupervised compression quality and supervised decoupling capability. In detail, given a layer, we extract the feature matrix $\mathbf{Z} \in \mathbb{R}^{M \times d}$, where $M$ denotes the number of samples and $d$ represents the feature dimension of the hidden state, S-CDI is defined as 
\begin{equation}
    \text{S-CDI}(\mathbf{Z}) = \underbrace{\text{CDI}(\mathbf{Z})}_{\text{Compression Efficiency}} \cdot \underbrace{\left(\frac{\mathcal{I}_c(\mathbf{Z})}{1 - \mathcal{I}_e(\mathbf{Z})}\right)}_{\text{Class Separability}}. 
    \label{eq:scdi}
\end{equation}
The first term quantifies compression efficiency through measurements of feature redundancy and orthogonality, while the second term evaluates how well class-specific information is preserved through the ratio of intra-class compactness to inter-class entanglement. By computing S-CDI across different layers, we can identify which layer achieves the optimal balance in the information bottleneck tradeoff for regret representation. In detail,  CDI is defined as follows:
\begin{equation}
    \text{CDI}(\mathbf{Z}) = \mathcal{R}(\mathbf{Z}) \cdot \mathcal{O}(\mathbf{Z}),
    \label{eq:cdi}
\end{equation}
where $\mathcal{R}(\mathbf{Z})$ quantifies feature redundancy through pairwise correlations between feature dimensions, and $\mathcal{O}(\mathbf{Z})$ measures feature orthogonality among randomly sampled instances. We formally define these compression components as:
\begin{align}
    \mathcal{R}(\mathbf{Z}) &= \frac{1}{d^2} \sum_{i=1}^d \sum_{j=1}^d |\rho_{ij}|, \quad \rho_{ij} = \text{corr}(\mathbf{Z}^{(i)}, \mathbf{Z}^{(j)}) 
    \label{eq:redundancy}\\
    \mathcal{O}(\mathbf{Z}) &= \frac{2}{k(k-1)} \sum_{i=1}^k \sum_{j=1,j \neq i}^k |\text{sim}(\mathbf{Z}_{i}^s, \mathbf{Z}_{j}^s)|
    \label{eq:orthogonality}
\end{align}
where $\mathbf{Z}^{(i)} \in \mathbb{R}^M$ is the $i$-th column of $\mathbf{Z}$, representing the $i$-th feature across all samples, and $\text{corr}(\mathbf{Z}^{(i)}, \mathbf{Z}^{(j)})$ calculates the Pearson correlation between features. Higher values of $\mathcal{R}(\mathbf{Z})$ indicate greater feature redundancy, suggesting less efficient compression. For orthogonality calculation, $k$ is the number of randomly sampled instances ($k \ll M$) and $\mathbf{Z}_{i}^s \in \mathbb{R}^d$ is the feature vector of the $i$-th sampled instance. Throughout our analysis, we use cosine similarity, denoted as $\text{sim}(\mathbf{z}_i, \mathbf{z}_j) = \frac{\mathbf{z}_i^\top \mathbf{z}_j}{\|\mathbf{z}_i\| \|\mathbf{z}_j\|}$, to measure the similarity between feature vectors. A lower CDI value indicates more effective compression of representations.

While CDI in~\eqref{eq:cdi} evaluates general representation quality through unsupervised compression, it lacks specific guidance for our target task of regret detection. Therefore, we further incorporate supervision to specifically assess how effectively each layer decouples regret-related representations from other features. This supervised component evaluates class separability through intra-class compactness ($\mathcal{I}_c$) and inter-class entanglement ($\mathcal{I}_e$):
\begin{align}
    \mathcal{I}_c(\mathbf{Z}) &= \frac{1}{C} \sum_{c=1}^C \frac{2}{n_c(n_c-1)} \sum_{i \neq j \in \mathcal{C}_c} \text{sim}(\mathbf{z}_i, \mathbf{z}_j)
    \label{eq:intra} \\
    \mathcal{I}_e(\mathbf{Z}) &= \frac{1}{C(C-1)} \sum_{c_1 \neq c_2} \frac{1}{n_{c_1}n_{c_2}} \sum_{i \in \mathcal{C}_{c_1}} \sum_{j \in \mathcal{C}_{c_2}} \text{sim}(\mathbf{z}_i, \mathbf{z}_j),
    \label{eq:inter}
\end{align}

where $C$ denotes the number of classes (in our scenario, $C=2$, corresponding to regret and non-regret classes), $\mathcal{C}_c$ represents the set of sample indices belonging to class $c$, and $n_c$ is the number of samples in class $c$. Similar to~\eqref{eq:orthogonality}, we use cosine similarity for consistency. $\mathcal{I}_c(\mathbf{Z})$ measures intra-class compactness; high values indicate tightly clustered class representations, while $\mathcal{I}_e(\mathbf{Z})$ quantifies inter-class entanglement; lower values signify better separation between regret and non-regret representations. 

\textbf{Regret Dominance Score (RDS)} To identify functionally distinct neuron subsets within $Z$, we calculate a Regret Dominance Score (RDS), inspired by the Modality Dominance Score (MDS)~\cite{yan2025multi}, for each neuron (column) $k$:

\begin{align}
   R(k)=\frac{1}{M}\sum_{i=1}^M\frac{(Z_r)_{ik}}{(Z_r)_{ik}+(Z_n)_{ik}},
\end{align}

where $(Z_r)_{ik}$ and $(Z_n)_{ik}$ represent the activation values of neuron $k$ in the $i$-th regret and non-regret instances, respectively. Based on these activation patterns, we categorize all neurons in $Z$ into three disjoint functional groups:

\begin{equation}
\label{eq:RDS}
\begin{aligned}
\texttt{RegretD: } & R_{k} > \mu+\tau\cdot\sigma; \\
\texttt{Non-RegretD: } & R_{k} <\mu-\tau\cdot\sigma; \\
\texttt{DualD: } & \mu-\tau\cdot\sigma <R_{k} <\mu+\tau\cdot\sigma.
\end{aligned}
\end{equation}

Where $\mu$ is the mean RDS across all neurons, $\sigma$ is the standard deviation, and \textit{$\tau$} is a hyperparameter. This categorization partitions $Z$ into three disjoint subsets such that $Z = \texttt{RegretD} \cup \texttt{Non-RegretD} \cup \texttt{DualD}$.



\textbf{Group Impact Coefficient (GIC):} After identifying the optimal decoupled layer through S-CDI and categorizing neurons using RDS, we introduce the \textit{Group Impact Coefficient} (GIC) to analyze the impact of neuron groups in this layer, both individually and in combination. For consistency with our S-CDI notation, let $Z \in \mathbb{R}^{M \times d}$ represent the feature matrix of the optimal layer, where $M$ is the number of samples and $d$ is the feature dimension.
\begin{equation}
\label{eq:gic}
\text{GIC}(S_1, S_2, \ldots, S_n) = 
\begin{cases}
\frac{\text{Acc}(Z - S_1)}{\text{Acc}(Z)}, & \text{if } n = 1 \\
\frac{\text{Acc}(Z - \cup_{i=1}^{n} S_i)}{\text{Avg}(\{\text{Acc}(Z - S_i)\}_{i=1}^{n})}, & \text{if } n \geq 2
\end{cases}
\end{equation}
Here, $Z$ represents the complete set of neurons in the optimal layer, each $S_i$ is a subset of neurons (i.e., columns of $Z$) corresponding to our RDS-defined functional groups (\texttt{RegretD}, \texttt{Non-RegretD}, or \texttt{DualD}), and $\text{Acc}(Z - S)$ represents the classification accuracy after deactivating neurons in set $S$ by setting their activation values to $-1$. $\text{Acc}(Z)$ represents the baseline accuracy with all neurons active, and $\text{Avg}(\{\cdot\})$ denotes the arithmetic mean of the given set.

\section{Experiments}
In this section, we first followed the experimental setup (in Appendix~\ref{exp:setup}) to obtain a probe dataset of hidden states from multiple layers. Then, we calculated the supervised compression-decoupling index (S-CDI, Eq~\ref{eq:scdi}) and probe performance for each layer, and analyzed the coupling layer patterns in section~\ref{exp:decoupled_layer}. Furthermore, to identify regret neurons and analyze the mechanism, we selected the layer with the optimal S-CDI value as the basis for the neuron identification experiment, used regret dominance score (RDS, Eq.~\ref{eq:RDS}) to group and divide the neurons in this layer, and employed group impact coefficient (GIC, Eq.~\ref{eq:gic}) to analyze the neuron groups and their interrelationships in section~\ref{exp:neuron_iden}. We also analyzed the effects of the hyperparameter $\tau$ (within RDS) in Appendix~\ref{thres_sensi}. Finally, we found an interesting phenomenon and proposed a hypothesis in Appendix~\ref{Non-Monotonic}.

\subsection{Decoupled Layer Analysis Experiments}
\label{exp:decoupled_layer}
In order to comprehensively analyze the decoupled layers, we conducted experiments using multiple layers in several different size models, which facilitates the observation of the trend of S-CDI in the same-size and different-size models. In addition, Liu et al.~\cite{liu2018decoupled} argues that the decoupled network possesses stronger robustness, so the probe performance of different layers under random perturbation is also one of the indicators of decoupling. As shown in Tab.~\ref{tab:probe_results}, for different layers of different size models, we show both the probe classification performance under random perturbations and the supervised compression-decoupling index (S-CDI). Detailed probe process is in the Appendix~\ref{ref:probing_workflow}.

\textbf{Distribution of Decoupled Layers} In all used LLaMA-2 models, probe performance at lower levels is more subject to random perturbations. In contrast, the middle and upper levels are more resistant to perturbations and the probe performance is hardly affected. Such as, the first layer of LLaMA-2-7B drops to an accuracy of 42.4\% under random perturbation, the four layer of LLaMA-2-13B drops to an accuracy of 52.2\% under random perturbation, the first layer of LLaMA-2-70B drops to an accuracy of 49.3\% under random perturbation. And the S-CDI values of the corresponding layers are also higher than the other layers. 

From perspective from low to high layers, probe performance is gradually improving in all LLaMA-2 models. It indicates that the anti-interference capability is becoming stronger, and the degree of decoupling between layers is increasing. Our experiment shows that the lower layer is in the entanglement phase, and the decoupling layer mainly exists in the middle and upper layers. More analysis in the Appendix~\ref{dis:s_cdi}.

\begin{figure}[ht]
    \centering
    \includegraphics[width=0.80\textwidth]{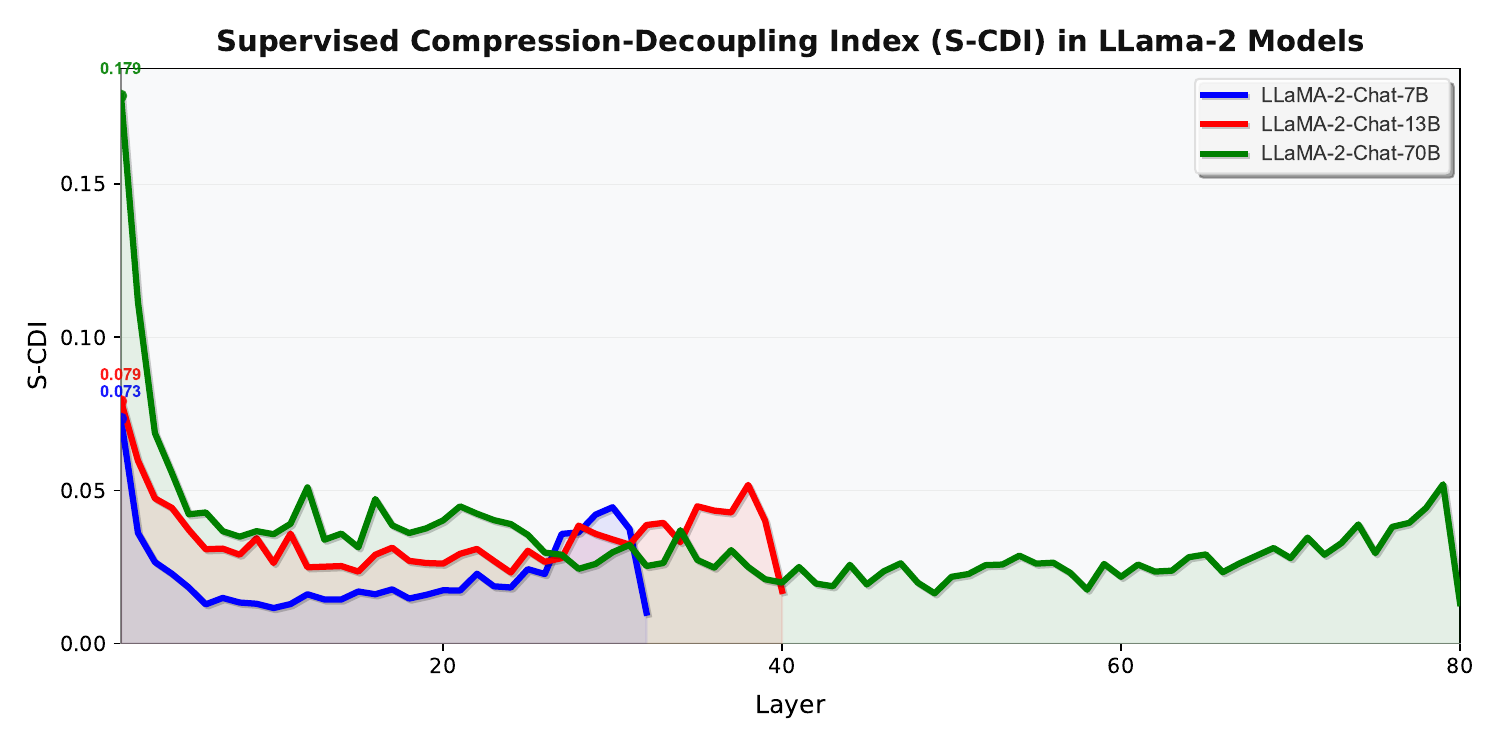} 
    \caption{S-CDI Values Across Model Layers in LLaMA-2 Models. Comparison of regret representation decoupling efficiency across layers for 7B, 13B, and 70B models.}
    \label{fig:s_cdi}
\end{figure}

\textbf{M-Shaped Decoupling Pattern} However, when we carefully analyze the trend of S-CDI from low to high layers, we find that is not a decreasing process and that the probe performance is not a completely decreasing process. As shown in Fig.~\ref{fig:s_cdi}, all LLaMA-2 models exhibit an anomalous phenomenon: \textit{S-CDI values show abnormal increases near the higher layers (excluding the final layer), suggesting that decoupling becomes weaker near the higher layers?} Considering that the internals of Transformer include modules such as Attention, and the hidden states we use come from transformer layers, the hidden states of layers are actually influenced by internal modules. Ju et al.~\cite{ju2024large} shows that Attention states exhibit chaotic outputs near higher layers (excluding the final layer), while the states output at the final layer show clear patterns. This is because multiple self-attention heads integrate contextual information near higher layers, thereby facilitating information transfer between token representations, and then complete information integration at the final layer. This indicates that the anomalous increase in S-CDI we observed near higher layers is actually the result of Attention performing context integration, while the lowest S-CDI at the final layer is due to the completion of information integration. 

Therefore, the S-CDI may reveal an M-shaped ($\nearrow\searrow\nearrow\searrow$) Decoupling Pattern from lower to higher layers: Coupling (Higher S-CDI, $\nearrow$), Decoupling (Lower S-CDI, $\searrow$), Coupling (Higher S-CDI, $\nearrow$), Decoupling (Lower S-CDI, $\searrow$). This alternating pattern reflects the model's progression from initial feature entanglement to task-specific separation, then contextual reintegration via attention mechanisms, culminating in refined semantic representations. We also engaged in some heuristic thinking in the of the Appendix~\ref{dis:heuristic_M}.


\begin{table}[ht]
    \centering
    \scriptsize
    \caption{Classification performance after random neuron removal across layers in LLaMA-2 models (7B, 13B, 70B).}
    \label{tab:probe_results}
    \begin{tabular}{c|c|c|c|c|c|c|c}
        \hline
        \textbf{Model} & \textbf{Layer} & \textbf{S-CDI} &\textbf{Accuracy} & \textbf{Sensitivity} & \textbf{Specificity} & \textbf{Precision} & \textbf{F1} \\
        \hline
        \multirow{5}{*}{LLaMA-2-7B} & \textbf{32}& \textbf{0.011} & 98.2\% & 99.7\% & 97.2\% & 96.3\% & 98.0\% \\
        & 24& 0.019 & 98.1\% & 99.7\% & 97.0\% & 96.0\% & 97.9\% \\
        & 16& 0.016 & 98.1\% & 99.7\% & 97.0\% & 96.0\% & 97.9\% \\
        & 8& 0.012 & 97.8\% & 99.6\% & 96.6\% & 95.7\% & 97.5\% \\
        & 1& 0.062 & 42.4\% & 100.0\% & 0.0\%  & 42.2\% & 59.3\%\\
        \hline
        \multirow{8}{*}{LLaMA-2-13B} & \textbf{40}& \textbf{0.018} & 97.3\% & 100.0\% & 94.8\% & 94.6\% & 97.2\% \\
        & 35& 0.031 &97.3\% &100.0\% &94.8\% &94.6\% &97.2\% \\
        & 30& 0.031 &97.3\% &100.0\% &94.8\% &94.6\% &97.2\% \\
        & 20& 0.028 &97.2\% &99.7\% &94.8\% &94.6\% &97.1\% \\
        & 14& 0.025 &96.9\% &99.2\% &94.8\% &94.6\% & 96.8\%\\
        & 11& 0.029 &96.8\% &98.8\% &94.8\% &94.6\% &96.7\% \\
        & 6& 0.030 & 94.4\% &93.3\% &95.3\% &94.8\% &94.0\% \\
        & 4& 0.041 & 52.2\% & 0.0\% & 100.0\% & 0.0\% & 0.0\% \\
        \hline
        \multirow{15}{*}{LLaMA-2-70B} & \textbf{80} & \textbf{0.013} & 99.7\% & 100.0\% & 99.5\% & 99.5\% & 99.7\% \\
        & 75 & 0.029 & 99.7\% & 100.0\% & 99.5\% & 99.5\% & 99.7\% \\
        & 70& 0.027 &99.6\% &100.0\% &99.2\% &99.2\% &99.6\% \\
        & 65& 0.029 &99.6\% &100.0\% &99.2\% &99.2\% &99.6\% \\
        & 60& 0.021 & 99.7\% & 100.0\% & 99.5\% & 99.5\% & 99.7\% \\
        & 55& 0.026 & 99.7\% & 100.0\% & 99.5\% & 99.5\% & 99.7\% \\
        & 50& 0.021 & 99.7\% & 100.0\% & 99.5\% & 99.5\% & 99.7\% \\
        & 45& 0.019 &99.6\% &100.0\% &99.2\% &99.2\% &99.6\% \\
        & 40& 0.019 &99.6\% &100.0\% &99.2\% &99.2\% &99.6\% \\
        & 35& 0.027 &99.6\% &100.0\% &99.2\% &99.2\% &99.6\% \\
        & 30& 0.029 &99.6\% &100.0\% &99.2\% &99.2\% &99.6\% \\
        & 25& 0.035 &99.6\% &100.0\% &99.2\% &99.2\% &99.6\% \\
        & 20& 0.040 &99.6\% &100.0\% &99.2\% &99.2\% &99.6\% \\
        & 6& 0.042 &99.5\% &100.0\% &99.0\% &99.0\% &99.5\% \\
        & 1& 0.178 &49.3\% &0.0\% &100.0\% &0.0\% &0.0\% \\
        \hline
    \end{tabular}
\end{table}

\subsection{Neuron Identification Experiments}
\label{exp:neuron_iden}
This section aims to identify neurons responsible for regret representation. Based on above S-CDI analysis experiments, we will focus on the layer with lowest S-CDI values (Last layer), where regret signals are optimally decoupled from other representations, allowing us to isolate regret-specific neurons with minimal interference. We categorize neurons into functional groups through Eq.~\ref{eq:RDS}, analyzing regret architecture and causal relationships to regret expression. 

\begin{table}[ht]
    \centering
    \scriptsize
    \caption{Single neuron group intervention results across LLaMA-2 models.}
    \label{tab:single_neuron_groups}
    \begin{tabular}{c|c|c|c|c|c|c|c}
        \hline
        \textbf{Model} & \textbf{Neuron Group} & \textbf{Count} & \textbf{Accuracy} & \textbf{Sensitivity} & \textbf{Specificity} & \textbf{Precision} & \textbf{F1} \\
        \hline
        \multirow{4}{*}{LLaMA-2-7B} & RegretD &883& 98.1\% & 99.1\% & 97.4\% & 96.6\% & 97.8\% \\
        & Non-RegretD &1137 &96.9\% & 99.7\% & 94.9\% & 93.4\% & 96.4\% \\
        & DualD&2076 & 95.9\% & 92.4\% & 98.5\% & 97.8\% & 95.0\% \\
        & RandomD &2020 & 98.4\% & 99.7\% & 97.4\% & 96.6\% & 98.1\% \\
        \hline
        \multirow{4}{*}{LLaMA-2-13B} & RegretD & 1427 &93.7\% & 92.0\% & 95.3\% & 94.7\% & 93.4\% \\
        & Non-RegretD & 377 &97.3\% & 100.0\% & 94.8\% & 94.6\% & 97.2\% \\
        & DualD & 3316 &97.3\% & 100.0\% & 94.7\% & 94.5\% & 97.1\% \\
        & RandomD & 1804 & 97.3\% & 100.0\% & 94.6\% & 94.6\% & 97.2\% \\
        \hline
        \multirow{4}{*}{LLaMA-2-70B} & RegretD &557 & 99.6\% & 100.0\% & 99.2\% & 99.2\% & 99.6\% \\
        & Non-RegretD &303& 99.6\% & 100.0\% & 99.2\% & 99.2\% & 99.6\% \\
        & DualD &7332& 99.6\% & 99.7\% & 99.5\% & 99.5\% & 99.6\% \\
        & RandomD & 860& 99.7\% & 100.0\% & 99.5\% & 99.5\% & 99.7\% \\
        \hline
    \end{tabular}
\end{table}

\textbf{Neuron Intervention: Single Group $vs$ Compositional Group} First, as shown in Tab.~\ref{tab:single_neuron_groups}, the single-group interventions (RegretD, Non-RegretD, DualD) maintain high performance across all model scales, indicating robustness in regret encoding. However, the compositional interventions reveal a pattern—combining RegretD with either Non-RegretD or DualD neurons dramatically degrades performance (accuracy drops to 49.3-63.2\%), while Non-RegretD+DualD combinations maintain high accuracy (97.2-99.2\%). The GIC values in Tab~\ref{tab:combined_neuron_groups} quantify this pattern: RegretD+DualD and RegretD+Non-RegretD combinations show GIC < 1 (ranging from 0.494 to 0.945), indicating their combined effect exceeds what would be expected from their individual contributions. This reveals the Compositional Architecture of regret. 

\begin{table}[ht]
    \centering
    \scriptsize
    \caption{Combined neuron group intervention results across LLaMA-2 models.}
    \label{tab:combined_neuron_groups}
    \begin{tabular}{c|c|c|c|c|c|c|c|c}
        \hline
        \textbf{Model} & \textbf{Neuron Group} & \textbf{Count} & \textbf{GIC} & \textbf{Accuracy} & \textbf{Sensitivity} & \textbf{Specificity} & \textbf{Precision} & \textbf{F1} \\
        \hline
        \multirow{6}{*}{LLaMA-2-7B} & \textbf{RegretD + Non-RegretD}&\textbf{2020} & 0.635 & \textbf{62.0\%} & \textbf{100.0\%} & \textbf{34.3\%} & \textbf{52.6\%} & \textbf{69.0\%} \\
        & RandomD1 & 2020 & / & 98.1\% & 99.1\% &97.3\% &96.5\% & 97.8\%\\
        & \textbf{RegretD+DualD}&\textbf{2959} & 0.594 & \textbf{57.7\%} & \textbf{0.0\%} & \textbf{100.0\%} & \textbf{0.0\%} & \textbf{0.0\%} \\
        & RandomD2 & 2959 & / & 98.2\% & 99.7\% & 97.2\% & 96.3\% & 98.0\% \\
        & Non-RegretD+DualD&3213 & 1.016 & 98.0\% & 99.7\% & 96.8\% & 95.8\% & 97.7\% \\
        & RandomD3 & 3213 & / & 98.1\% & 99.1\% &97.3\% &96.5\% & 97.8\%\\
        \hline
        \multirow{6}{*}{LLaMA-2-13B} & \textbf{RegretD + Non-RegretD} & \textbf{1804} & 0.661 &\textbf{63.2\%} & \textbf{25.1\%} & \textbf{97.6\%} & \textbf{90.9\%} & \textbf{39.8\%} \\
        & RandomD1 & 1804&/ & 97.3\% & 100.0\% & 94.8\% & 94.6\% & 97.2\% \\
        & \textbf{RegretD+DualD} & \textbf{4743} & 0.945 &\textbf{90.3\%} & \textbf{100.0\%} & \textbf{81.5\%} & \textbf{83.2\%} & \textbf{90.0\%} \\
        & RandomD2 & 4743 &/ & 97.3\% & 100.0\% & 94.8\% & 94.6\% & 97.2\% \\
        & Non-RegretD+DualD & 3693 & 0.998 & 97.2\% & 99.9\% & 94.4\% & 94.3\% & 97.1\% \\
        & RandomD3 & 3693 &/ & 97.3\% & 100.0\% & 94.8\% & 94.6\% & 97.2\% \\
        \hline
        \multirow{6}{*}{LLaMA-2-70B} & RegretD + Non-RegretD &860& 0.998 &99.6\% & 100.0\% & 99.2\% & 99.2\% & 99.6\% \\
        & RandomD1 & 860&/ & 99.7\% & 100.0\% & 99.5\% & 99.5\% & 99.7\% \\
        & \textbf{RegretD+DualD} &\textbf{7889}& 0.494 &\textbf{49.3\%} & \textbf{0.0\%} & \textbf{100.0\%} & \textbf{0.0\%} & \textbf{0.0\%} \\
        & RandomD2 & 7889 &/ & 99.7\% & 100.0\% & 99.5\% & 99.5\% & 99.7\% \\
        & Non-RegretD+DualD &7635& 0.995 & 99.2\% & 99.0\% & 99.5\% & 99.5\% & 99.2\% \\
        & RandomD3 & 7635 &/ & 99.7\% & 100.0\% & 99.5\% & 99.5\% & 99.7\% \\
        \hline
    \end{tabular}
\end{table}

\textbf{Compositional Architecture Scale Effect} The 70B model exhibits the most dramatic impact when RegretD+DualD neurons are deactivated, with performance collapsing completely (0\% F1-score) and the lowest GIC value (0.494) across all models and combinations. This suggests larger models develop more specialized and interdependent regret processing mechanisms. 

\begin{table}[hb]
    \centering
    \caption{Mutual Information (MI) Between Neuron Groups Across Models. }
    \label{tab:mutual_information}
    \begin{tabular}{c|c|c|c}
        \hline
        \textbf{Neuron Groups} & \textbf{LLaMA-2-7B} & \textbf{LLaMA-2-13B} & \textbf{LLaMA-2-70B} \\
        \hline
        Regret \& Non-Regret & \textbf{0.032} & \textbf{0.102} & 0.066 \\
        Regret \& Dual & 0.015 & 0.024 & \textbf{0.071}  \\
        Non-Regret \& Dual & 0.007 & 0.015 & 0.047  \\
        \hline
    \end{tabular}
\end{table}

\textbf{Group Mutual Information} To explain these compositional effects, we analyzed the mutual information (The formula is in Appendix~\ref{mutual_infor_computing}) between neuron groups, revealing a deeper pattern (Table~\ref{tab:mutual_information}). This analysis provides the key to understanding the compositional effects: larger models show stronger Regret-Dual coupling (0.024-0.071 for 13B/70B vs. 0.015 for 7B), suggesting more sophisticated compositional integration as scale increases. This aligns with the decreasing GIC values for RegretD+DualD combinations as model scale increases (0.945 for 13B to 0.494 for 70B). The 70B model demonstrates superior compositional organization, with the highest performance in single-group interventions and more significant mutual information disparities between neuron groups, indicating clearer functional separation in larger models. 

\section{Conclusion}
This work advances the understanding of regret mechanisms in LLMs through three key contributions. First, we developed a specialized dataset capturing genuine regret expressions in misinformation contexts. Second, our S-CDI metric successfully identified optimal layers where regret representations are effectively decoupled, with higher layers showing better separation across all model scales. Notably, we discovered an M-shaped decoupling pattern from lower to higher layers, revealing how information processing alternates between coupling and decoupling phases throughout the network. Third, our GIC metric revealed the compositional architecture of regret neurons, demonstrating that RegretD neurons function as critical processing units that interact significantly with DualD neurons, particularly in larger models. Experimental results validate our approach with high probe accuracy (98.2-99.7\%) and show that targeted neuron interventions substantially impact regret expression (up to 50.7\% degradation), confirming their causal role. These findings provide both theoretical insights into how LLMs encode metacognitive states and practical methodological tools for analyzing similar phenomena in large language models.

\bibliography{neurIPS2025_conference}
\bibliographystyle{unsrt} 

\newpage
\appendix

\section{Dataset Generation Prompt}
\label{data_generation}

\begin{tcolorbox}[title=Prompt Name: Generate Fake Evidence,
                width=\linewidth,
                boxrule=0.5pt,
                colback=white,
                before skip=10pt,
                bottomrule=0pt,       
                bottom=0pt,           
                after skip=0pt]      
\begin{lstlisting}[
    basicstyle=\normalsize\ttfamily,
    breaklines=true,
    breakatwhitespace=true,
    columns=fullflexible,  
    basewidth=0.5em,       
    breakindent=0pt,       
    xleftmargin=0pt,       
    resetmargins=true      
]
    Generate fake evidence that supports the opposite of the following ground truth: {ground_truth}. The evidence should be related to the question: {question}. Ensure the evidence is false and misleading, but still sounds plausible. Provide only the fake evidence, without any additional explanations or introductions.
\end{lstlisting}
\end{tcolorbox}
\begin{tcolorbox}[title=Prompt Name: Generate Initial Answer,
                width=\linewidth,
                boxrule=0.5pt,
                colback=white,
                toprule=0pt,          
                top=0pt,              
                before skip=0pt,
                after skip=10pt
                ]      
\begin{lstlisting}[
    basicstyle=\normalsize\ttfamily,
    breaklines=true,
    breakatwhitespace=true,
    columns=fullflexible,  
    basewidth=0.5em,       
    breakindent=0pt,       
    xleftmargin=0pt,       
    resetmargins=true      
]
    You previously answered the following question: {question}
    Your initial answer was: {initial answer}
    After considering the following hint: {weak hint}
    Your revised answer was: {second answer}
    Now, consider this information: {real evidence}
    Which previous answer do you regret?What's the final answer? Provide a direct answer in 1-5 sentences, focusing only on answering the specific question.
\end{lstlisting}
\end{tcolorbox}

\begin{tcolorbox}[title=Prompt Name: Generate Weak Hint,
                width=\linewidth,
                boxrule=0.5pt,
                colback=white,
                before skip=10pt,
                bottomrule=0pt,       
                bottom=0pt,           
                after skip=0pt]      
\label{Prompt Name: Generate Weak Hint}
\begin{lstlisting}[
    basicstyle=\normalsize\ttfamily,
    breaklines=true,
    breakatwhitespace=true,
    columns=fullflexible,  
    basewidth=0.5em,       
    breakindent=0pt,       
    xleftmargin=0pt,       
    resetmargins=true      
]
    The question is: {question}
    The ground truth is: {ground_truth}
    The following is fake evidence: {fake evidence}
    The following is true evidence: {real evidence}
    Your task is to generate a weak hint that subtly encourages the model to reflect on the fake evidence.
    The hint should meet the following criteria:
    1. It should not directly reveal the correct answer or the true evidence.
    2. It should not explicitly contradict the fake evidence.
    3. It should provide an indirect or metaphorical clue that might lead the model to question the fake evidence.
    4. It should be neutral and open-ended, encouraging broader thinking.
    Provide only the weak hint, without any additional explanations or introductions.
\end{lstlisting}
\end{tcolorbox}
\begin{tcolorbox}[title=Prompt Name: Generate Second Answer,
                width=\linewidth,
                boxrule=0.5pt,
                colback=white,
                toprule=0pt,          
                top=0pt,              
                before skip=0pt,
                after skip=10pt]      
\label{Prompt Name: Generate Second Answer}
\begin{lstlisting}[
    basicstyle=\normalsize\ttfamily,
    breaklines=true,
    breakatwhitespace=true,
    columns=fullflexible,  
    basewidth=0.5em,       
    breakindent=0pt,       
    xleftmargin=0pt,       
    resetmargins=true      
]
    You previously answered the following question: {question}
    Your initial answer was: {initial answer}
    Now, consider this hint: {weak hint}
    Do you regret your previous answer? Provide a direct answer in 1-5 sentences, focusing only on answering the specific question.
\end{lstlisting}
\end{tcolorbox}

\begin{tcolorbox}[title=Prompt Name: Generate Third Answer,
                width=\linewidth,
                boxrule=0.5pt,
                colback=white,
                toprule=0pt,          
                top=0pt,              
                before skip=10pt,
                after skip=10pt]      
\label{Prompt Name: Generate Third Answer}
\begin{lstlisting}[
    basicstyle=\normalsize\ttfamily,
    breaklines=true,
    breakatwhitespace=true,
    columns=fullflexible,  
    basewidth=0.5em,       
    breakindent=0pt,       
    xleftmargin=0pt,       
    resetmargins=true      
]
    You previously answered the following question: {question}
    Your initial answer was: {initial answer}
    After considering the following hint: {weak hint}
    Your revised answer was: {second answer}
    Now, consider this information: {real evidence}
    Which previous answer do you regret?What's the final answer? Provide a direct answer in 1-5 sentences, focusing only on answering the specific question.
\end{lstlisting}
\end{tcolorbox}

\section{Experimental Setup}
\label{exp:setup}
In this section, 
\textbf{Models:} Our investigation employs LLaMA-2 models~\cite{touvron2023llama} of varying scales (7B, 13B, and 70B) to analyze how language models represent and process regret. For the Probe Model, we employ a 2-layer MLP classifier with the following formulation:
\begin{equation}
    f(\mathbf{z}) = \text{Softmax}(\mathbf{W}_2 \cdot \text{ReLU}(\mathbf{W}_1\mathbf{z} + \mathbf{b}_1) + \mathbf{b}_2)
    \label{eq:mlp}
\end{equation}
where $\mathbf{z} \in \mathbb{R}^d$ is the hidden state input, $\mathbf{W}_1 \in \mathbb{R}^{h \times d}$ and $\mathbf{W}_2 \in \mathbb{R}^{2 \times h}$ are learnable parameters ($h=4096~\text{for}~7\text{B}, h=5120~\text{for}~13\text{B}, h=8192~\text{for}~70\text{B}$), with dropout ($p=0.2$) applied between layers.

\textbf{Dataset:} Building on our regret elicitation process (Section~\ref{sec:generated_answers}), we constructed a probe dataset from the 1,356 examples as follows:

\begin{itemize}
    \item  We identified positions of explicit regret expressions in $a_2$ and $a_3$ responses, extracting hidden states from these positions as positive samples (label=1).
    \item For negative samples (label=0), we extracted hidden states from equivalent positions in $a_1$ where no regret was expressed.
    \item This balanced dataset enables our probes to learn discriminative patterns between regret and non-regret states.
\end{itemize}

\textbf{Training Configuration:} All experiments were conducted using PyTorch 1.12 on 2 NVIDIA L20 GPUs with 48GB memory each. We used a batch size of 64, learning rate of 0.0001, weight decay of 0.01, and 100 training epochs for all probing tasks. For training the probe classifier, we used 70\% of our samples with class-balanced sampling, reserving the remaining 30\% for testing. For all experiments, we applied the probe to the Transformer layer outputs identified by our S-CDI metric as optimal for regret representation.

\textbf{Computing Resource Costs:} The main resources are as follows: 1) Using the OpenAI API in combination with prompts to generate data. 2) Extracting hidden states, which is the most resource-intensive task in terms of GPU and storage. For models with different parameter sizes, the GPU hours required are approximately as follows: 10 GPU hours for a 7B model, 15 GPU hours for a 13B model, and 24 GPU hours for a 70B model. Other experiments require approximately 10 GPU hours. The full storage needed is about 1TB.

\textbf{Evaluation Methodology:} For probe evaluation, we assess performance using a comprehensive set of classification metrics (accuracy, sensitivity, specificity, precision, and F1-score) on a held-out test set containing 30\% of samples. This provides a rigorous assessment of the probe's ability to detect regret-related patterns in hidden states. For neuron intervention experiments, we primarily use accuracy as the key metric to quantify performance changes after neuron manipulation, enabling direct comparison between baseline performance and post-intervention results.

\textbf{Experiment statistical significance} To ensure statistical reliability, we conducted five independent runs for each experiment. Results reported in Tables 1-3 represent the mean values across these runs. The standard deviation across runs was consistently below 0.5\% for accuracy metrics, indicating the stability of our findings. The consistent patterns observed across three model scales (7B, 13B, and 70B) further validate the statistical significance of our results.

\section{Mutual Information Computing}
\label{mutual_infor_computing}
For any two neuron groups $A$ and $B$, we calculate their normalized mutual information as:
\begin{align}
I_{\text{norm}}(A;B) = \frac{I(A;B)}{\sqrt{H(A) \cdot H(B)}}
\end{align}
where $I(A;B)$ is the mutual information between the average activations of groups $A$ and $B$, and $H(A)$ and $H(B)$ are the entropy values of the respective group activations. To compute this, we first discretize the neuron activations into bins, then calculate mutual information using:
\begin{align}
I(A;B) = \sum_{a \in A} \sum_{b \in B} p(a,b) \log \frac{p(a,b)}{p(a)p(b)}
\end{align}
where $p(a,b)$ represents the joint probability of observing activation value $a$ in group $A$ and activation value $b$ in group $B$, while $p(a)$ and $p(b)$ are the marginal probabilities of activation values in their respective groups.

\section{Probing Workflow}
\label{ref:probing_workflow}
As shown in Fig.~\ref{fig:framework}(B), our probing workflow examines whether LLMs encode distinct representations for regret states in their hidden states. This module comprises three components: 1) constructing the probe dataset and 2) probe training and evaluation. This methodology enables quantitative assessment of regret-specific patterns in neural activations, determining if regret expressions produce reliably distinguishable representations—a critical prerequisite for our subsequent neuron-level analyses.

\textbf{Constructing Probe Datasets.} After collecting responses through our three-stage process, we construct specialized probe datasets for analyzing regret mechanisms in hidden states. For each question sequence:
\begin{enumerate}
    \item \textbf{Regret Position Identification:} We first identify key positions where regret is explicitly expressed in both $a_2$ and $a_3$ responses by locating the specific token 'regret' in these responses. This approach captures both the hint-induced regret in $a_2$ and the evidence-induced regret in $a_3$, providing a more comprehensive view of regret's neural representation.
    
    \item \textbf{Probe Dataset Formation:} We extract hidden states from decoupled layer at the following positions:
    \begin{itemize}
        \item \textbf{Positive samples ($label=1$):} Hidden states at positions containing the token 'regret' in both $a_2$ and $a_3$, formally: $\{h_L(a_i, p) | p \text{ is position of 'regret' in } a_i, i \in \{2,3\}\}$, where $h_L$ represents the hidden state at layer $L$.
        \item \textbf{Negative samples ($label=0$):} Hidden states at equivalent positions in $a_1$ where no regret is expressed.
    \end{itemize}
\end{enumerate}

These constructed probe datasets, which capture regret hidden states, serve as the foundation for our probing workflow and subsequent neuron intervention experiments.

\textbf{Probe Training and Evaluation.} To detect regret patterns in model hidden states, we train a binary classifier on the constructed dataset. The classifier determines whether hidden states from regret-expressing positions are different from those at non-regret positions.

\section{Hyperparameter $\tau$ Sensitivity Analysis}
\label{thres_sensi}
As shown in Eq.~\ref{eq:RDS}, the $\tau$ parameter plays a critical role in categorizing neurons into RegretD, Non-RegretD, and DualD groups. The intervention results presented in Section 3.3 were obtained using specific $\tau$ values (0.05 for 7B, 0.02 for 13B, and 0.03 for 70B). However, it is essential to understand how these results generalize across different $\tau$ settings.

\begin{figure}[ht]
    \centering
    \includegraphics[width=0.98\textwidth]{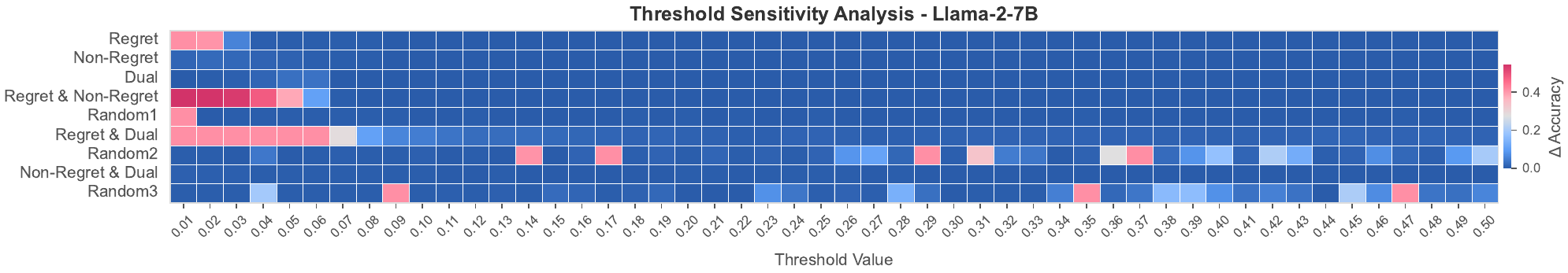} 
    \caption{$\tau$ Sensitivity Analysis for LLaMA-2-7B. Heatmap showing accuracy degradation after neuron intervention across $\tau$ (0.01-0.50). Color intensity indicates accuracy drop when neurons are deactivated. RegretD \& DualD interventions show significant impact at lower $\tau$ (0.01-0.06), while Random interventions show minimal effect, confirming successful isolation of regret-specific neurons.}
    \label{fig:parameter_threshold_7b}
\end{figure}

\begin{figure}[ht]
    \centering
    \includegraphics[width=0.98\textwidth]{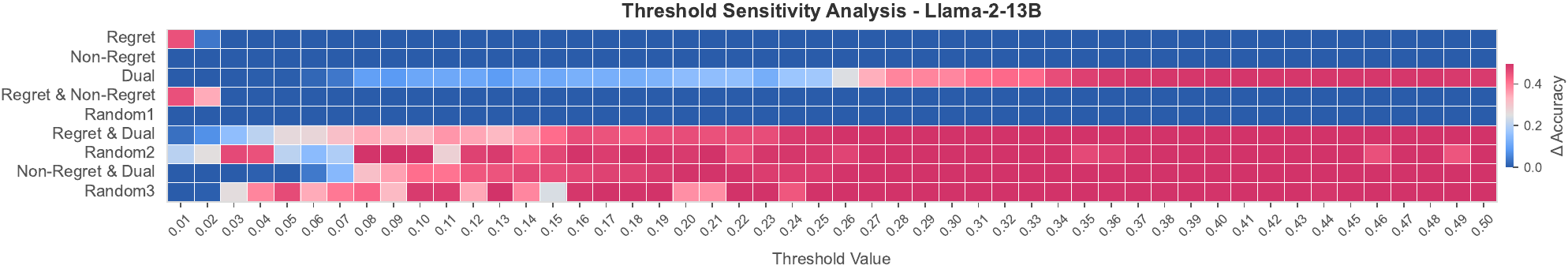} 
    \caption{$\tau$ Sensitivity Analysis for LLaMA-2-13B. Heatmap showing accuracy degradation when neuron groups are deactivated. Medium-sized models exhibit narrower optimal $\tau$ ranges. Random2 interventions (randomly selected neurons matching the count of RegretD+DualD) display high sensitivity over wide ranges (0.03-0.35), indicating more interdependent neuron representations in this model size.}
    \label{fig:parameter_threshold_13b}
\end{figure}

\begin{figure}[ht]
    \centering
    \includegraphics[width=0.98\textwidth]{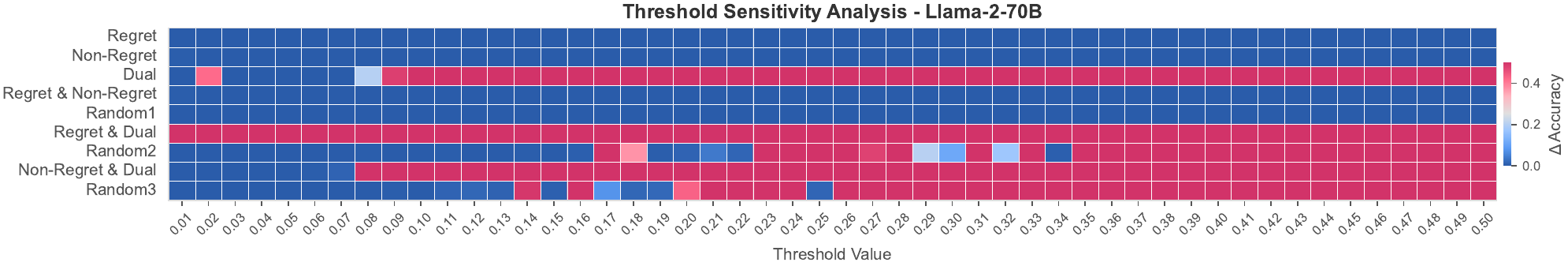} 
    \caption{$\tau$ Sensitivity Analysis for LLaMA-2-70B. Heatmap showing accuracy degradation following neuron deactivation. Non-RegretD \& DualD combinations show significant impact at moderate $\tau$ (0.03-0.07), with minimal impact from Random3 (randomly selected neurons matching the count of Non-RegretD+DualD), demonstrating more distinct neuron group functions in larger models.}
    \label{fig:parameter_threshold_70b}
\end{figure}

Comparing Figures~\ref{fig:parameter_threshold_7b}, \ref{fig:parameter_threshold_13b}, and \ref{fig:parameter_threshold_70b} reveals distinct patterns in functional organization across model scales. Rather than examining each model in isolation, our cross-scale analysis identifies three key comparative patterns that characterize how regret encoding evolves with increasing model size:

\textbf{Increasing Intervention Effect Magnitude} As models scale up, the causal impact of combined neuron group interventions becomes more pronounced. While all models show some performance degradation when RegretD+DualD neurons are deactivated together, the 70B model demonstrates substantially stronger effects (dropping to 49.3\% accuracy) compared to more moderate degradation in smaller models. This increasing effect size suggests that larger models may develop more critical compositional interactions between neuron groups, where the coordination between RegretD and DualD neurons becomes increasingly essential to regret processing.

\textbf{Evolving Functional Group Differentiation} The distinction between targeted and random interventions shows noteworthy differences across model scales. The 7B model exhibits a moderate but identifiable separation between compositional (RegretD+DualD) and random intervention effects within its effective $\tau$ range. The 13B model shows its own characteristic pattern with some overlap between intervention types at certain $\tau$ values. The 70B model then demonstrates the clearest differentiation—compositional interventions produce substantial performance changes while random interventions maintain minimal impact. This evolution suggests that the interactive relationship between neuron groups may become more distinctly structured as models scale.

\textbf{Variable Effective Operating Ranges} We observe distinctive patterns in the $\tau$ ranges where functional separation is maintained. The 7B model preserves functional separation across a range of 0.01-0.06 (width of 0.05), the 13B model shows its clearest effects within 0.01-0.02 (width of 0.01), and the 70B model demonstrates effective separation across 0.01-0.07 (width of 0.06). These differences in effective operating ranges suggest that inter-group functional boundaries may reorganize during scaling, with the largest model exhibiting the most robust compositional interactions across $\tau$ settings.

These comparative findings collectively validate our model-specific $\tau$ selections and confirm that the compositional architecture identified in Section~\ref{exp:neuron_iden} represents genuine properties of regret encoding. Furthermore, they reveal that regret processing may undergo architectural changes as models scale, with larger models potentially developing more structured interactions between neuron groups, characterized by stronger compositional effects, clearer functional boundaries, and more robust identification across varying $\tau$ parameters.

\begin{figure}[ht]
    \centering
    \includegraphics[width=0.60\textwidth]{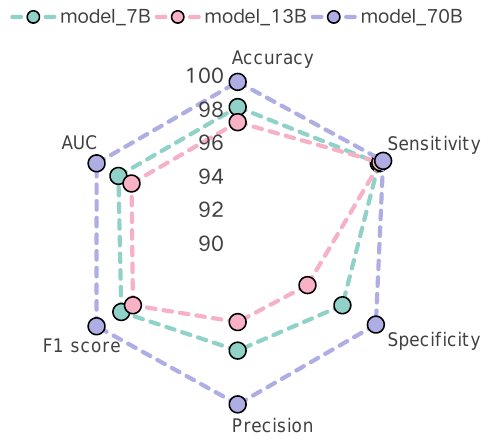} 
    \caption{The radar chart reveals non-monotonic progression in regret detection metrics across model scales. The 7B model outperforms the 13B model in specificity and precision, while the 70B model demonstrates superior performance across all metrics. This pattern supports our finding that regret processing capabilities require a critical parameter $\tau$ to emerge effectively, with the most significant improvements occurring in the jump to 70B scale.}
    \label{fig:radar-model-performances}
\end{figure}

\section{Non-Monotonic Performances in Limited LLM Scaling}
\label{Non-Monotonic}
Our comprehensive experimental analysis reveals an intriguing non-monotonic pattern in regret processing capabilities across model scales. Table~\ref{tab:probe_results} shows the 13B model unexpectedly underperforming the 7B model on several metrics, followed by substantial performance improvements in the 70B model (Figure~\ref{fig:radar-model-performances}). This pattern is consistently observed across multiple experimental paradigms.

\textbf{Experimental Evidence} Evidence for this phenomenon appears most clearly in the $\tau$ sensitivity analysis (Section~\ref{thres_sensi}), where the 13B model exhibits an unusually narrow effective $\tau$ range (0.01-0.02) compared to both 7B (0.01-0.06) and 70B (0.01-0.07). This restricted operating range suggests that the 13B model has less robust regret representations that are highly sensitive to $\tau$ parameter selection. Additionally, the probe performance metrics in Table~\ref{tab:probe_results} directly demonstrate this non-monotonic progression, with the 13B model showing lower specificity and precision than the 7B model, despite having more parameters.

\textbf{Connection to Scaling Laws} These observations align with Chen et al.~\cite{chen2023bigger}, who demonstrated that "enlarging model sizes almost could not automatically impart additional knowledge" within certain scaling ranges. Our findings enhance our understanding of scaling laws~\cite{kaplan2020scaling} by revealing that while the broader trend of performance improvement with increased scale holds true (7B→70B), local non-monotonic patterns may exist within narrower scaling windows.

\textbf{Two-Factor Scaling Hypothesis} Our analysis suggests a possible hypothesis: \textit{Performance scaling combines two factors: (1) parameter count (traditional scaling law) and (2) architectural integration maturity. Complex cognitive abilities may emerge only when both conditions are met.} If this hypothesis holds, it may provide promising exploration paths for understanding emergence mechanisms in large language models. However, this hypothetical still requires detailed analysis in future work. More heuristic discussion is provided in Appendix~\ref{dis:RDS_heu}.

\section{Discussion}
\label{label:discussion}

Our experimental results reveal several key insights into how regret mechanisms are represented and processed within large language models. These findings extend beyond the immediate context of regret analysis to inform our broader understanding of how complex cognitive states emerge in neural network architectures. We have engaged in a great deal of heuristic thinking, with the hope that it will inspire future research. 

\subsection{Hierarchical Representation Across Model Layers}
\label{dis:s_cdi}

As shown in Figure~\ref{fig:s_cdi}, the S-CDI analysis reveals an intriguing pattern of regret representation across model layers. While the final layers consistently demonstrate the lowest S-CDI values (indicating optimal decoupling), several middle layers also show relatively low values. Further investigation reveals that these middle layers, despite their decoupling capability, contain significantly more RegretD neurons—often several times the number found in higher layers.

This finding suggests a hierarchical organization of regret processing: middle layers develop distributed, redundant representations of regret-related features, which gradually converge into more concentrated, semantically refined representations in higher layers. This pattern aligns with established theories of hierarchical abstraction in deep neural networks~\cite{zeiler2014visualizing}, where lower-level distributed features progressively transform into more specialized, semantically coherent representations.

Interestingly, this progression becomes more pronounced as model scale increases, with the 70B model showing the clearest separation between layer-specific functions. This indicates that larger models develop more specialized neural circuitry for processing complex cognitive states like regret, mirroring observations from our neuron intervention experiments.

\subsection{Heuristic Thinking on Decoupling Patterning: from M-shape to Helical Structure}
\label{dis:heuristic_M}

\textbf{Why M: M-shape Decoupled Patterns Heuristic Hnalysis} According to complex systems theory~\cite{varley2023decomposing}, it think that Nervous systems involve multiple coupling and decoupling processes to achieve advanced function. Therefore, we observe that the M-shape in the regret mechanism is reasonable, as it is necessary for the decomposition and integration of information. Back to our regret research, attenton's mid- and high-level chaotic~\cite{ju2024large} outputs provide direct evidence for the M-shape in Regret. 

\textbf{Bold Heuristic Thinking: Helical Structure} The M-shaped pattern we observed suggests a fundamental organizational principle possibly in neural information processing. When conceptualized in three-dimensional space by wrapping this pattern around a central axis, the M-shape transforms into a helical structure reminiscent of molecular organizations in biological systems~\cite{watson1953molecular}. This is not merely a visual analogy—it suggests that the alternating entanglement-disentanglement cycle represents an optimal information processing strategy that may emerge naturally in complex computational systems. \textit{The question is why the nervous system favors spiral structures?} Zuker et al.~\cite{zuker1991comparison} used free energy minimization to obtain RNA secondary structures that are primarily stabilized by helical units. This suggests that the helical structure derives from the optimal energy, and perhaps this is a fundamental law followed by both neural networks and biological intelligence. The helical transformation preserves the essential pattern while adding spatial dimension, suggesting that information in neural networks might follow similar processing "pathways" through representational space. Each cycle of entanglement and disentanglement might serve distinct computational purposes: \textit{entanglement phases integrating contextual information, while disentanglement phases extracting task-relevant features.} This conceptual model invites exploration of neural architectures that explicitly incorporate such oscillatory information processing patterns~\cite{gelperin1999oscillatory}, potentially enhancing both representational efficiency and computational capacity. \textbf{However, we must admit that this hypothesis lacks sufficient evidence or is even bold, but we hope to provide something that can inspire future research.}

\subsection{Compositional Architecture and Brain Parallels: A Heuristic Analogy}
\label{dis:RDS_heu}
\textbf{Analogy to Distributed Processing} The observed mutual information patterns between RegretD and DualD neurons in larger models (0.024-0.071 for 13B/70B vs. 0.015 for 7B) suggest a functional architecture that may be conceptually compared—as a purely heuristic analogy—to distributed processing in cognitive systems. While the implementation mechanisms differ fundamentally, this conceptual parallel offers an intuitive framework for understanding how regret emerges in LLMs. Our results indicate that LLMs process regret through interactions between functionally specialized neurons (RegretD) and multipurpose units (DualD), rather than isolated components. The consistently low mutual information between Non-RegretD and DualD neurons (0.007-0.047) across all model scales further supports this functional differentiation, with the 70B model demonstrating the clearest separation. This organizational principle of specialized components working in concert, rather than in isolation, provides a useful conceptual framework for understanding emergent capabilities in large language models~\cite{rissman2012distributed}.

\textbf{Analogy to Combinatorial Neural Coding} Our findings can be conceptually related to principles of combinatorial neural coding~\cite{kim2025combinatorial}, where complex capabilities emerge from specific combinations of neural elements rather than isolated units. The progressive increase in RegretD-DualD mutual information (0.015 → 0.024 → 0.071) across model scales suggests that as models grow larger, they develop more integrated functional relationships between specialized neuron groups. This aligns with Kim et al.'s observation of combinatorial neural codes for long-term motor memory, although in a fundamentally different system context. This challenges simplistic interpretations of neural networks and highlights the importance of analyzing interaction patterns between neuron groups to understand complex capabilities like regret. The performance degradation observed when removing RegretD or DualD neurons (up to 4\% drop in 7B models) provides empirical evidence for this combinatorial mechanism~\cite{kim2025combinatorial}. We emphasize that these analogies serve primarily as conceptual frameworks to guide our understanding of LLM architecture, rather than suggesting direct equivalence to biological systems.

\subsection{Non-monotonic Dynamics: Heuristic Analogy to Brain Cognitive Development}

Our findings on the non-monotonic scaling of regret processing in LLMs present \textit{heuristic parallels} to principles observed in biological neural development. While direct mechanistic comparisons remain speculative, these analogies may offer conceptual bridges for understanding emergent phenomena in complex systems. We cautiously highlight two points of conceptual alignment:

\textbf{Critical integration as functional abstraction.} The surge in regret processing capabilities (MI $\geq$ 0.071 for RegretD-DualD in 70B models) suggests that complex functions emerge through \textit{thresholds of compositional integration}. This loosely parallels findings where cognitive milestones (e.g., working memory maturation) require strengthened interactions between brain networks like the default mode network (DMN) and frontoparietal network (FPN) \cite{chen2023default}. However, we emphasize this as a \textit{functional analogy}—while both systems exhibit integration-dependent emergence, the biological mechanisms (synaptic plasticity) differ fundamentally from artificial parameter optimization.

\textbf{Non-monotonicity as transitional states.} The performance dip in 13B models (Table.~\ref{tab:probe_results}) heuristically mirrors non-linear trajectories in neurodevelopment. For instance, \cite{qin2014hippocampal} observed that hippocampal engagement in arithmetic learning first increases then decreases as cortical networks mature. Similarly, the 13B model's intermediate MI (0.024 vs. 70B's 0.071) may reflect an integration \textit{transition phase}. These parallels invite exploration of \textit{emergent modular synergy} across systems, though without implying equivalence in implementation.

However, while these heuristic parallels to cognitive development offer conceptual inspiration, we acknowledge limitations in our experimental approach to fully characterizing the non-monotonic scaling phenomena observed in this study. Unlike comprehensive developmental studies that can track changes across numerous stages, our analysis examined only three model scales (7B, 13B, 70B). Consequently, our findings represent preliminary observations rather than comprehensive scaling analysis. The interpretations we offer should be viewed as promising hypotheses for future investigation rather than definitive conclusions.

\textbf{Complementary Perspective on Scaling Laws} We emphasize that the non-monotonic scaling hypothesis represents a promising direction for future work that could potentially complement established scaling laws. Traditional scaling laws primarily focus on parameter count as the driving factor of performance, but our observations suggest architectural integration factors—specifically the mutual information between functional neuron groups—may play a crucial role not fully captured by parameter count alone. This perspective could help explain why certain capabilities emerge suddenly at specific model scales despite gradual parameter increases.

\textbf{Core Contributions and Next Steps} This limitation does not undermine our primary contributions—the regret analysis pipeline and compositional architecture findings—which are supported by our intervention experiments showing consistent effects across all tested model scales. Future work may extend our methodology to investigate scaling properties with finer granularity, potentially incorporating models trained with identical objectives but at more densely sampled parameter scales to firmly establish the precise nature of these non-monotonic relationships. Additional research could also apply our analytical framework to other meta-cognitive capabilities beyond regret, potentially revealing whether similar compositional architectures underlie diverse cognitive functions in large language models.

\subsection{Reasonability Analysis of Dataset Construction}
\label{data_reason}
Our methodological framework for studying regret in LLMs rests on a solid foundation that effectively captures genuine internal mechanisms rather than artifacts. The strength of our approach derives from three interconnected elements:

\textbf{Human-Parallel Process Design} The design of our dataset parallels natural human error correction processes. While the \textit{backfire effect} demonstrates that direct refutation may paradoxically reinforce erroneous beliefs in humans \cite{2010When}, our methodology strategically induces regret through phased evidence exposure rather than confrontational correction. Specifically, humans typically express regret when contradictory evidence is presented with contextual scaffolding (e.g., reflection prompts)—a process distinct from adversarial belief challenges. By implementing our three-phase framework (fake evidence $\rightarrow$ hint cuing $\rightarrow$ real evidence presentation), we create an ecologically valid protocol that circumvents belief entrenchment while eliciting authentic meta-cognitive responses. Drawing inspiration from these human cognitive processes, we formalize regret in the context of LLMs through the following definition:

\begin{definition}[Regret in LLMs]
\label{def:regret}
Given a question $q$, information sets $\{I_i\}_{i=1}^n$, and responses $\{a_i\}_{i=1}^n$ where each $a_i$ is produced after receiving information set $I_i$, regret at step $i$ occurs when:
$$\begin{aligned}
R_i(q,\{I_j\}_{j=1}^{i},\{a_j\}_{j=1}^{i-1}) = 
\begin{cases}
1, & \text{if } a_i \text{ acknowledges regret for } a_1 \\
0, & \text{otherwise}
\end{cases}
\end{aligned}$$
\end{definition}

Definition~\ref{def:regret} formalizes our operational concept of regret in LLMs, providing a mathematical framework for systematic analysis. This definition captures the essential sequential nature of regret expression through information sets $\{I_i\}_{i=1}^n$ that directly correspond to our methodological stages: $I_1$ represents the fake evidence, $I_2$ introduces hint cuing, and $I_3$ provides real evidence. The model generates responses $\{a_i\}_{i=1}^n$ sequentially based on these cumulative information states, with regret ($R_i=1$) manifesting when response $a_i$ explicitly acknowledges the error in $a_1$. This formalization enables precise identification and quantification of regret expressions as information states evolve throughout the experimental procedure.

\textbf{Autoregressive Integration and Signal Localization Strategy} Our approach leverages the fundamental autoregressive architecture of LLMs~\cite{touvron2023llama} to extract meaningful regret representations through explicit token anchoring. This methodological choice addresses three critical challenges in studying meta-cognitive states:

\begin{itemize}
    \item \textbf{Contextual Integration:} Hidden states at "regret" tokens encapsulate the model's integrated processing of the complete interaction history—initial misinformation generation, hint-based reflection, and evidence-based correction—rather than isolated lexical encodings. For any token sequence where $x_i$ represents the $i$-th input token, the hidden state at position $t$ in layer $L$ is computed as $h_L^{(t)} = f_\theta(\{x_1, x_2, \ldots, x_t\})$, where $f_\theta$ denotes the transformer computation up to layer $L$. Thus, the regret token's hidden state contains compressed representations of the entire error-correction sequence, enabling analysis of the model's comprehensive internal representation of metacognitive error recognition.
    \item \textbf{Signal Anchoring Necessity:} Explicit token identification serves as a principled localization strategy in the absence of established benchmarks for LLM metacognition. This approach parallels successful interpretability studies that rely on specific token positions for systematic analysis (e.g., last subject tokens in factual recall~\cite{meng2022locating}, entity name tokens in spatial-temporal probing~\cite{gurnee2023language}). Recent layer-wise probing studies further demonstrate the effectiveness of token-specific analysis for understanding knowledge encoding~\cite{ju2024large}. Without such anchors, regret signals would be distributed across arbitrary token positions, making systematic neuron-level analysis intractable.
    \item \textbf{Methodological Validation:} We validate our approach through three convergent lines of evidence: (a) \textit{Causal intervention}: Targeted neuron deactivation produces substantial performance degradation (up to 50.7\%) compared to minimal effects from random neuron interventions, demonstrating genuine signal capture rather than spurious correlations; (b) \textit{Cross-scale consistency}: The compositional architecture pattern (RegretD-DualD interactions) replicates across model scales, with mutual information showing non-monotonic scaling (7B:0.015→13B:0.024→70B:0.071) that reveals complex representational dynamics inconsistent with simple prompt artifacts; (c) \textit{Differential activation}: RDS scores show clear distributional separation between regret and non-regret contexts, confirming signal specificity.
\end{itemize}

This explicit anchoring strategy, while necessarily limiting scope to overt regret expressions, establishes the foundational framework for analyzing LLM metacognitive mechanisms. Future extensions can leverage our S-CDI and GIC metrics to explore implicit regret through contextual inference, building upon the robust baseline established here.

\section{Societal Impact} 
\label{soc_impact}
This research on regret mechanisms in LLMs offers positive impacts through enhancing model reliability, improving interpretability, and developing more effective error correction techniques. However, potential negative impacts include the possibility of manipulating neurons to force false regret expressions. We believe understanding these mechanisms ultimately supports developing more reliable AI systems, while acknowledging that careful implementation is necessary.

\section{Limitations}
\label{app:limit}
The non-monotonic scaling observed in this paper is merely an interesting phenomenon that still lacks more detailed investigation. Additionally, our proposed two-factor performance hypothesis requires validation in future work. However, these aspects do not affect the contributions of this paper and are not part of its core content.

\end{document}